\documentclass[10pt, conference]{IEEEtran}
\ifCLASSINFOpdf
\else
\fi
\usepackage{cite}
\usepackage{amstext,amsmath,amssymb,amsfonts}
\usepackage{algorithm}
\usepackage{algorithmic}
\usepackage{verbatim}
\usepackage{mathtools}
\usepackage{url}
\usepackage{graphicx}
\usepackage{subfigure}
\usepackage{hyperref}
\usepackage{paralist}
\usepackage{textcomp}
\usepackage{xcolor}
\usepackage{balance}
\def\BibTeX{{\rm B\kern-.05em{\sc i\kern-.025em b}\kern-.08em
    T\kern-.1667em\lower.7ex\hbox{E}\kern-.125emX}}

\definecolor{goldenyellow}{rgb}{1.0, 0.8, 0.06}
\definecolor{kellygreen}{rgb}{0.3, 0.73, 0.09}
\definecolor{electricpurple}{rgb}{0.75, 0.0, 1.0}
\definecolor{OliveGreen}{rgb}{0, 0.6, 0}
\definecolor{RoyalBlue}{rgb}{0.25, 0.41, 0.88}

\begin{document}
\title{An LSTM model for Twitter Sentiment Analysis}


\author{\IEEEauthorblockN{Md Parvez Mollah}
\IEEEauthorblockA{University of New Mexico\\
Albuquerque, USA\\
parvez@unm.edu}
}


\maketitle

\begin{abstract}
Sentiment analysis on social media such as Twitter provides organizations and individuals an effective way to monitor public emotions towards them and their competitors. As a result, sentiment analysis has become an important and challenging task. In this work, we have collected seven publicly available and manually annotated twitter sentiment datasets. We create a new training and testing dataset from the collected datasets. We develop an LSTM model to classify sentiment of a tweet and evaluate the model with the new dataset.
\end{abstract}


% \fi
\IEEEpeerreviewmaketitle

\section{Introduction}\label{sec:intro}
With the advancement of technology, social media platforms have become an integral part of human lives. People express and share their thoughts and opinions on all kinds of topics and events on social media platforms. Twitter is one of the most popular social networking platforms. It allows people to post messages to share their interests, favorites, opinions, and sentiments towards various topics and issues they encounter in their daily life. The messages are called tweets, which are real-time and at most $280$ characters.

Twitter provides us access to the unprompted views of a wide set of users on particular products or events. The opinions or expressions of sentiment about organizations, products, and events have proven extremely useful for marketing \cite{jiang2011} and social studies \cite{thelwall2010}. As a result, twitter sentiment analysis has become a widely popular research topic.

Most of the works in twitter sentiment classification focus on obtaining sentiment features by analyzing lexical and syntactic features \cite{saif2012,jianqiang2015,thelwall2011,montejo2014}. These features are expressed explicitly through sentiment words, exclamation marks, emoticons etc. Relying on these explicit features can lead to incorrect results. For example, people sometimes use ":)" emoticon sarcastically when they feel offended. Classifying a tweet as positive sentiment based on ":)" emoticon can be incorrect in such cases.

Some machine learning models have been proposed for sentiment classification in literature \cite{jianqiang2018,sulaiman2019}. They use datasets from various sources to train and test their models. However, the models are trained and tested for each dataset separately, which can induce overfitting problem due to prior bias in the test set.

In this work, we have collected seven publicly available twitter sentiment datatsets, which are manually annotated. We construct a new training and testing dataset by combining these seven datasets. We develop a Long Short-Term Memory (LSTM) networks model for the twitter sentiment classification task and evaluate the model with our newly created training and testing dataset.

\section{Data}\label{sec:datasets}

\begin{table*}[t]
    \centering
    \caption{Summary of the seven datasets}
    \begin{tabular}{lcccccccc}
    \hline
    
    \hline
      Dataset & Src & \#Tweets & \#Positive & \#Negative & \#Neutral & \#Other & \#Irrelevant\\
      \hline
STS-Test & \cite{go2009} & 498 & 182 & 177 & 139 & -- & --\\
HCR & \cite{speriosu2011} & 2515 & 541 & 1381 & 470 & 45 & 79\\
OMD & \cite{shamma2008} & 3259 & 1606 & 845 & 289 & 519 & --\\
SS-Twitter & \cite{thelwall2011} & 4242 & 507 & 297 & 3438 & -- & --\\
Sanders & \cite{sanders} & 5113 & 519 & 572 & 2333 & 1689 & --\\
SemEval & \cite{nakov2013} & 9395 & 3460 & 1441 & 2062 & 2432 & --\\
STS-Gold & \cite{saif2013} & 2034 & 632 & 1402 & -- & -- & --\\
    \hline
    
    \hline
    \end{tabular}
    \label{tab:datasum}
\end{table*}

In this section, we describe the collected datasets and the data preprocessing steps.

\subsection{Datasets description}
We have gathered seven different datasets that are widely used in the twitter sentiment analysis. The main characteristic of the datasets is that they are manually annotated, providing a reliable set of judgements over the tweets. Tweets in these datasets have been
annotated with different sentiment labels including positive, negative, neutral, mixed, other and irrelevant. Table~\ref{tab:datasum} displays the distribution of tweets in the
seven collected datasets according to these sentiment labels.

\subsection{Data preprocessing}

\begin{figure}[htb]
    \centering
    \includegraphics[scale=0.42]{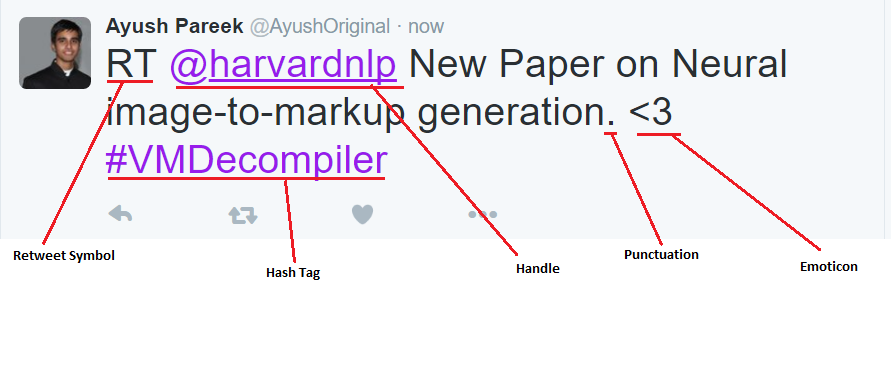}
    \vspace{-0.3cm}
    \caption{Various features seen in tweets \cite{ayush2018}.}
    \label{fig:tweet_features}
\end{figure}
Contents posted by users on the internet are hardly usable due to their different characteristics. It is a necessary steps to normalize the text by applying a series of preprocessing steps. We have applied a set of preprocessing steps to decrease the size of the features set to make it usable for learning algorithms. Figure \ref{fig:tweet_features} shows various features seen in tweets on Twitter. Next, we present a brief description of preprocessing steps taken.

\subsubsection{Hashtags}
A hashtag is a word or an no-spaced phrase prefixed with the hash symbol (\#). These are used to both naming subjects and phrases that are currently in trending topics. For example, \#Ashes2021, \#SpiderManNoWayHome etc. We replace the hashtags with HASH\_$\backslash$1 expression.

\subsubsection{Handles}
Every Twitter user has a unique username. Any thing directed towards that user can be indicated be writing their username preceded by ‘@’. Thus, these are like proper nouns. For example, @Cristiano, @rogerfederar etc. Handles are replaced by the expression HNDL\_$\backslash$1.

\subsubsection{URLs}
Users often share hyperlinks in their tweets. Twitter shortens them using its in-house URL shortening service, like http://t.co/FCWXoUd8 - such links also enables Twitter to alert users if the link leads out of its domain. From the point of view of sentiment classification, a particular URL is not important. However, presence of a URL can be an important feature. We replace URLs by the keyword URL.

\subsubsection{Emoticons}
In social media websites, emoticons are used heavily by the users. We identify different types of emoticons and substitute them by one of these suitable keywords EMOT\_SMILEY, EMOT\_LAUGH, EMOT\_LOVE, EMOT\_WINK, EMOT\_FROWN, EMOT\_CRY.

\subsubsection{Punctuations}
Though not all punctuations are important from the point of view of sentiment classification but some of these, like question mark, exclamation mark can also provide information about the sentiments of the text. We replace every word boundary by a list of relevant punctuations present at that point.

\subsubsection{Constuct combined dataset}
Our collected datasets contain tweets with various sentiment labels. We only consider the tweets that have been annotated as positive, negative, and neutral, and remove the tweets which have been marked as irrelevant and other. We have used the preprocessing code provided by \cite{ayush2018} to normalize the tweet data. We merge all the preprocessed tweets and their labels into a single dataset. We randomy shuffle the dataset and divide it into two sets: training (80\%) and testing (20\%).

\section{Proposed Model}\label{sec:proposed_model}

\begin{figure}[htb]
    \centering
    \includegraphics[scale=0.42]{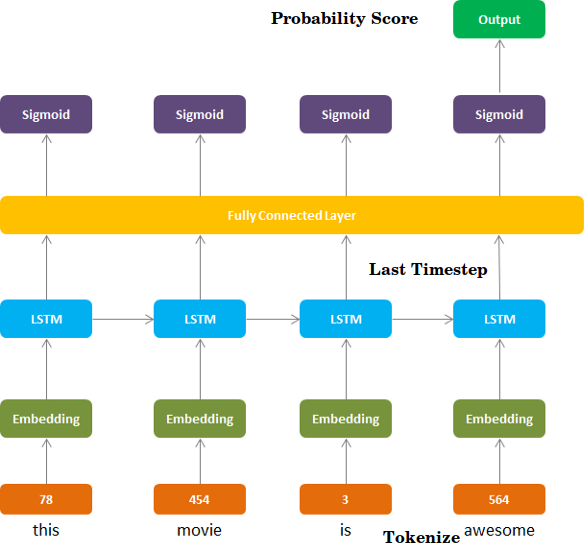}
    \caption{Proposed LSTM mode architecture \cite{samarth2019}.}
    \label{fig:model_arch}
\end{figure}

We have implemented an Long Short-Term Memory (LSTM) networks model. Figure \ref{fig:model_arch} illustrates the architecture of the proposed LSTM model. The layers of the model are as follows \cite{samarth2019}:
\begin{enumerate}
  \item Tokenize: This is not a layer for LSTM network but a mandatory step of converting the words into tokens (integers).
  \item Embedding Layer: This layer converts the word tokens (integers) into embedding of specific size. We use GloVe \cite{pennington2014glove} word embeddings to convert the words into fix-sized vectors.
  \item LSTM Layer: This layer is defined by the number of neurons, number of hidden layers, activation functions in the layers etc.
  \item Fully Connected Layer: It maps the output of LSTM layer to a desired output size.
  \item Sigmoid Activation Layer: It turns all output values in a value between 0 and 1.
  \item Output: Sigmoid output from the last timestep is considered as the final output of the network.
\end{enumerate}

\section{Experimental Results}\label{sec:expresults}

\begin{table}[h]
\fontsize{8pt}{10pt}
\selectfont
\centering
\caption{LSTM Model Parameters}\label{tab:LSTMModelParameters}
\begin{tabular}{|l | l|}
\hline
\hline
    Parameter & Value \\ \hline
    Number of layers & 1 \\ \hline
    Number of neuron in hidden layer & 100\\ \hline  	
    Activation function in hidden layer & relu\\\hline 
	Number of neuron in output layer & 3 \\\hline
	Activation function in output layer & sigmoid\\\hline
	Loss function & categorical\_crossentropy\\\hline
	Optimizer & adam\\\hline
	Number of epochs & 50\\\hline
	Batch size & 256\\\hline
	metrics & accuracy\\\hline
 	\hline
\end{tabular}
\end{table}

In the experimental evaluation, we compare the accuracy of our model with the popular VADER (Valence Aware Dictionary and sEntiment Reasoner) \cite{vader}, which is a lexicon and rule-based sentiment analysis tool, that is specifically attuned to sentiments expressed in social media. All the parameters  used in our LSTM model are specified in Table \ref{tab:LSTMModelParameters}.

\subsection{Accuracy comparison}

\begin{table}[h]
\fontsize{8pt}{10pt}
\selectfont
\centering
\caption{Tweets distribution in test set}\label{tab:tweet_dist}
\begin{tabular}{|l | l|}
\hline
\hline
    Tweet label & Number of tweets\\ \hline
    Total & 4459 \\ \hline
    Positive & 1473 \\ \hline
    Negative & 1231 \\ \hline
    Neutral & 1755 \\ \hline
 	\hline
\end{tabular}
\end{table}

\begin{table}[h]
\fontsize{8pt}{10pt}
\selectfont
\centering
\caption{Accuracy comparison}\label{tab:accuracy_comparison}
\begin{tabular}{|l | l| l| l| l|}
\hline
\hline
    Model & Overall accuracy & Pos accuracy & Neg accuracy & Neu accuracy \\ \hline
    LSTM & \textbf{0.68} & \textbf{0.77} & \textbf{0.68} & \textbf{0.61} \\ \hline
    VADER & 0.51 & 0.66 & 0.44 & 0.43 \\ \hline  	
 	\hline
\end{tabular}
\end{table}

Table \ref{tab:tweet_dist} shows the distribution of total positive, negative, and neutral tweets in the test set. We observe that the test set is almost balanced. Table \ref{tab:accuracy_comparison} presents the accuracy comparison between our LSTM model and the VADER model. We notice that the overall accuracy of our proposed LSTM model is much higher than the overall accuracy of VADER. It is interesting to note that both of the models shows higher accuracy in identifying positive tweets than the negative and neutral tweets.

\subsection{Runtime comparison}
Our proposed LSTM model takes 10 minutes to train and classify the tweets. This time includes the conversion of words to vectors. With the pre-trained model, prediction takes only 10 seconds to run. On the other hand, the VADER model runs in 3 seconds. Though the VADER model runs faster than our LSTM model, the poor accuracy of VADER makes it less useful.

\section{Discussion}
In this work, we have shown that machine learning models perform better than lexicon and rule-based algorithm for the sentiment classification task. Though the accuracy of our proposed LSTM model is not up to the mark, the model can be tweaked in several ways to achieve better results. For example, we have used GloVe for word embeddings, which does not contain vector representation of emoticons. So, our model could not focus on emoticons in the text. Using a different embeddings containing emoticon vectorizations might lead to a better accuracy. If we were doing it all over again, we would use a different embeddings. Also, we would use attention mechanism to focus on emoticons and punctuations so that the model could perform better.

%\section{Conclusion}
%Since I have done this project alone, I did all the work.

%\balance
\bibliographystyle{IEEEtran}
\bibliography{bare_conf}

\end{document}